\documentclass{article}

     \usepackage[preprint,nonatbib]{neurips_2020}

\usepackage[utf8]{inputenc} 
\usepackage[T1]{fontenc}    
\usepackage{hyperref}       
\usepackage{url}            
\usepackage{booktabs}       
\usepackage{amsfonts}       
\usepackage{nicefrac}       
\usepackage{microtype}      
\usepackage{amsmath}        
\usepackage{graphicx}       
\usepackage{subfigure}      
\usepackage{wrapfig}        
\usepackage{multirow}
\usepackage{array}

\title{PVSNet: Pixelwise Visibility-Aware Multi-View Stereo Network}

\author{%
  Qingshan Xu and Wenbing Tao\thanks{Corresponding author} \\
  National Key Laboratory of Science and Technology on Multispectral Information Processing\\
  School of Artifical Intelligence and Automation\\
  Huazhong University of Science and Technology, China \\
  \texttt{\{qingshanxu, wenbingtao\}@hust.edu.cn} \\
}

\begin{document}

\maketitle

\begin{abstract}
Recently, learning-based multi-view stereo methods have achieved promising results. However, they all overlook the visibility difference among different views, which leads to an indiscriminate multi-view similarity definition and greatly limits their performance on datasets with strong viewpoint variations. In this paper, a Pixelwise Visibility-aware multi-view Stereo Network (PVSNet) is proposed for robust dense 3D reconstruction. We present a pixelwise visibility network to learn the visibility information for different neighboring images before computing the multi-view similarity, and then construct an adaptive weighted cost volume with the visibility information. 
Moreover, we present an anti-noise training strategy that introduces disturbing views during model training to make the pixelwise visibility network more distinguishable to unrelated views, which 
is different with the existing learning methods that only use two best neighboring views for training. To the best of our knowledge, PVSNet is the first deep learning framework that is able to capture the visibility information of different neighboring views. In this way, our method can be generalized well to different types of datasets, especially the ETH3D high-res benchmark with strong viewpoint variations. Extensive experiments show that PVSNet achieves the state-of-the-art performance on different datasets.  
\end{abstract}

\section{Introduction}

Multi-view stereo (MVS) aims to recover 3D geometry of a scene from a set of calibrated images and is a fundamental problem in computer vision. Despite of the abundant relevant literature~\cite{Seitz2006Comparison,Furukawa2015MST,Furukawa2010Accurate,Schonberger2016Pixelwise,Yao2018MVSNet,Xu2019Multi,Xu2020Planar}, it still faces many challenges due to a variety of real-world problems, such as occlusions, illumination and unstructured viewing geometry.

\begin{figure}[t]
	\setlength{\abovecaptionskip}{0pt}
	\setlength{\belowcaptionskip}{0pt}
	\centering
	\includegraphics[width=0.96\textwidth]{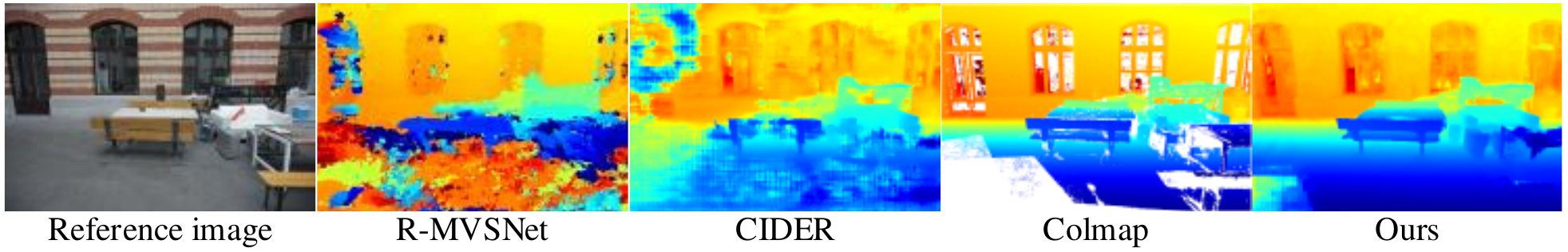}
	\caption{Depth map results using R-MVSNet~\cite{Yao2019RMVSNet}, CIDER~\cite{Xu2020Learning}, Colmap~\cite{Schonberger2016Pixelwise} and Ours on ETH3D high-res multi-view benchmark~\cite{Schops2017Multi}.}
	\label{introduction}
\end{figure}

Recently, learning-based multi-view stereo methods have achieved promising results ~\cite{Yao2018MVSNet,Yao2019RMVSNet,Chen2019Point,Xu2020Learning,Cheng2020UCSNet} and even outperform some traditional methods, e.g., Colmap~\cite{Schonberger2016Pixelwise} on Tanks and Temples dataset~\cite{Knapitsch2017TTB}. However, what is puzzling is that they have never been evaluated on ETH3D high-res benchmark~\cite{Schops2017Multi}. To clarify this issue, we choose two recently published learning-based methods to test on ETH3D high-res benchmark. The results in Fig.~\ref{introduction} demonstrate that their estimated depth maps seriously degrade compared with Colmap. Why does this happen? To explain this phenomenon, we first analyze the difference in characteristics of these two datasets. Tanks and Temples dataset provides video sequences as input, and the viewpoint between adjacent images often changes slightly. This means that almost all areas of some neighboring images are visible in the reference image. Unlike the Tanks and Temples dataset, the images provided by ETH3D high-res benchmark contain strong variations in viewpoint, resulting in complicated visibility association. This requires MVS methods to consider the pixelwise visibility information of different neighboring images. On the other hand, the existing learning-based methods~\cite{Yao2018MVSNet,Yao2019RMVSNet,Chen2019Point,Xu2020Learning,Cheng2020UCSNet} are tailored for video sequences with continuous viewpoint changes. Since they assume that there exist neighboring images that have strong visibility association with the reference image, they usually select these images as input from the perspective of global view selection and treat these images equally to construct an indiscriminate multi-view aggregated cost volume. Therefore, visibility estimation is totally ignored in these networks. However, treating each neighboring image equally will make the cost volume susceptible to the noise from unrelated neighboring images. 
This greatly limits the performance of learning-based methods on datasets like ETH3D high-res benchmark with strong variations in viewpoint. 
In contrast, some traditional methods like Colmap~\cite{Schonberger2016Pixelwise} and ACMM~\cite{Xu2019Multi} consider the pixelwise visibility information of neighboring images in their multi-view similarity computation. In this way, they achieve robust estimation on different types of datasets, including ETH3D high-res benchmark.
However, in real-world scenarios, especially in large-scale 3D reconstruction, the viewpoint of input images usually changes greatly.
\emph{Therefore, to make learning-based methods truly feasible in practice, it is an urgent problem to learn the pixelwise visibility information of neighboring images in deep networks.} 
In fact, the visibility estimation, which is validated in our subsequent experiments, is not only necessary for the datasets with strong viewpoint changes like the ETH3D high-res benchmark, but also crucial for the datasets with video sequences as input like the Tanks and Temples dataset. 

In this work, we propose Pixelwise Visibility-aware multi-view Stereo Network (PVSNet) to achieve robust 3D reconstruction on different datasets. Our network first learns the pixelwise visibility information of each neighboring image with respect to the reference image. Then multiple pairwise costs are aggregated into a unified representation by using this visibility information. Specifically, through plane-sweeping with multiple sampling depths, we construct a two-view cost volume for each neighboring image. Based on the two-view cost volume, by regressing a 2D visibility probabilistic map for each neighboring image, we explicitly establish the visibility association between each neighboring image and the reference image. With the help of visibility information, the previous two-view cost volumes can be aggregated into a robust unified one in a weighted manner. This greatly reduces the influence of noise from unrelated neighboring images. 
To make the pixelwise visibility network more distinguishable to unrelated views, we further propose an anti-noise training strategy (AN) to introduce disturbing views during model training while the existing learning methods only use two best neighboring views for training. This can greatly improve the robustness of the network.

To summarize, our main contributions are as follows: 
\textbf{1)} We propose Pixelwise Visibility-aware multi-view Stereo Network (PVSNet) for robust multi-view depth estimation. To the best of our knowledge, PVSNet is the first deep learning framework that is able to capture the visibility information of neighboring images and can be truly applied to datasets with strong viewpoint changes. 
\textbf{2)} We propose a way to regress 2D visibility maps from two-view cost volumes. The visibility maps can reflect the influence of occlusion, illumination, and unstructured viewing geometry. This allows good views to occupy more weights in the final cost volume representation. 
\textbf{3)} We present a new training strategy that introduces disturbing views to improve the robustness of our pixelwise visibility network. 
Extensive experiments have been conducted to elaborate on the superiority of our proposed pixelwise visibility-aware multi-view similarity measure. We demonstrate our novel PVSNet achieves state-of-the-art performance on various datasets.

\section{Related work}

Our work focuses on the visibility estimation of neighboring views to boost the performance of learning-based MVS methods. We briefly review the related techniques in the literature below.

\textbf{Traditional multi-view stereo.} Some traditional methods rely on robust pairwise similarity statistics to compute multi-view similarity. Kang et al.~\cite{Kang2001Handling} propose to include the best $50\%$ of all pairwise similarities. Subsequently, Galliani et al.~\cite{Galliani2015Massively} slightly change this constraint to select the best $K$  neighboring views. On the other hand, some approaches~\cite{Goesele2006Multi,Furukawa2010Accurate} introduce a certain threshold to exclude extremely bad neighboring views. 
However, since the above metrics do not explicitly model the visibility information, they cannot account for the importance of different neighboring views. 

To explicitly evaluate the visibility of different neighboring views, some iterative methods are proposed. Zheng et al.~\cite{Zheng2014PatchMatch} design a probabilistic graphical model to jointly estimate visibility and geometry. Based on this framework, Sch\"{o}nberger et al.~\cite{Schonberger2016Pixelwise} simultaneously consider a variety of photometric and geometric priors to improve the robustness and accuracy of geometry and visibility estimation. Xu and Tao~\cite{Xu2019Multi} leverage a multi-hypothesis joint voting scheme to determine a consistent set of selected views for different depth candidates. These iterative methods reflect the difference of neighboring views considering occlusion, illumination and viewing geometry, which inspires our work to model the visibility information in learning-based multi-view stereo methods.   

\textbf{Learning-based multi-view stereo.} 
Hartmann et al.~\cite{Hartmann2017Learned} propose to learn a multi-patch matching function by n-way Siamese networks and the mean operation. SurfaceNet~\cite{Ji2017Surfacenet} encodes camera parameters implicitly by mapping image appearance into 3D voxels and predicts if voxels are on the geometry surface. To select credible view pairs, it crops two representative patches around the projected center voxel and learns the relative weight for different image pairs from the perspective of global view selection. LSM~\cite{Kar2017Learning} unprojects image features into 3D feature grids by perspective geometry and fuses them into a unified one by recurrent neural networks. It implicitly considers the importance of different views but may be sensitive to the ordering of input images. To make networks order-agnostic and adapt to an arbitrary number of input images, DeepMVS~\cite{Huang2018DeepMVS} utilizes a max-pooling layer to select the most useful view information but cannot make full use of multi-view information. MVSNet~\cite{Yao2018MVSNet} and many following works~\cite{Xue2019MVSCRF,Yao2019RMVSNet,Chen2019Point} adopt a variance-based multi-view similarity metric to capture the second moment information. 
In essence, this metric treats each neighboring view equally and cannot distinguish useful views. Similarly, the mean operation used in~\cite{Luo2019P,Xu2020Learning,Im2019Dpsnet} also faces the above problem.

\section{Method}

\begin{figure*}[t]
	\setlength{\abovecaptionskip}{0pt}
	\setlength{\belowcaptionskip}{0pt}
	\centering
	\includegraphics[width=0.92\textwidth]{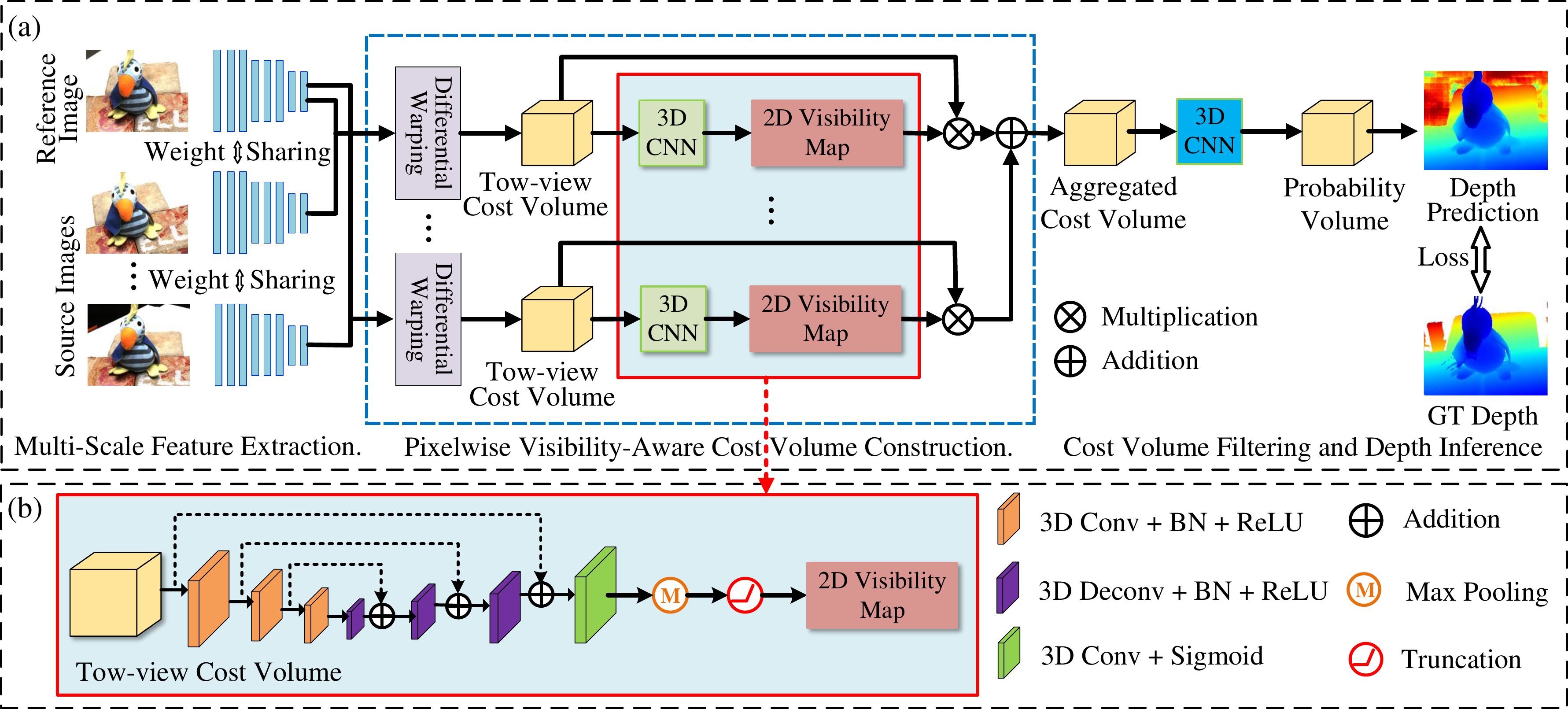}
	\caption{(a) shows the overview of our proposed PVSNet. (b) shows the network architecture of the pixelwise visibility network, which corresponds to the red box of (a).}
	\label{network}
\end{figure*}



Given a reference image $I_\text{ref}=I_{0}$ and source images $\boldsymbol{I}_\text{src}=\{I_{i}|i=1,\cdots, N-1\}$, our proposed network aims at inferring the depth map for the reference image when their camera parameters are known, where $N$ denotes the total number of input images. The overall architecture of PVSNet is depicted in Fig.~\ref{network} (a). Multi-scale image features are first extracted with a shared feature extraction module for all input images. Then the two-view cost
volume between each source image and the reference image is constructed by differential warping. In contrast to most existing methods~\cite{Yao2018MVSNet,Xu2020Learning,Im2019Dpsnet} that directly perform mean operation or variance calculation on multiple two-view cost volumes, the key novelty of our method is to estimate the pixelwise visibility information of different source images before computing the aggregated cost volume (cf. the red box in Fig.~\ref{network} (a)). Based on the visibility information, the two-view cost volumes are aggregated into a weighted one. Finally, the depth map of the reference image is predicted by cost volume filtering and depth inference.

\subsection{Multi-scale feature extraction}

To make the multi-view similarity measure robust to illumination and context-aware for ambiguous regions, our feature extraction employs a small image encoder to convert the raw RGB information into multi-scale deep image features. Following MVSNet~\cite{Yao2018MVSNet}, we apply an eight-layer 2D CNN to downsample the original image size ($3\times H \times W$) to $F \times \frac{H}{4} \times \frac{W}{4}$, where $H$ and $W$ denote the image height and width respectively, and $F=32$ is the number of feature channels.   

\subsection{Pixelwise visibility-aware cost volume construction}

The key of our framework is to learn pixelwise visibility maps for each source image. To this end, we first construct the two-view cost volume between each source image and the reference image. We then regress the visibility map of each source image based on each cost volume. With these visibility maps, multiple two-view cost volumes are further aggregated into a unified cost volume.

\textbf{Two-view cost volume construction.} 
Inspired by the traditional plane sweeping stereo~\cite{Collins1996Space}, we construct two-view cost volumes by warping the extracted deep features of each source image into the camera coordinate of the reference image based on multiple sampled depth hypotheses. Specifically, for a pixel ${\bf p}$ in the reference image $I_\text{ref}$, its corresponding projected point in the source image $I_i$ is 
\begin{equation}
{\bf p}_{i,j}={\bf K}_{i}({\bf R }_{\text{ref},i}({\bf K}_\text{ref}^{-1}{\bf p}d_{j})+{\bf t}_{\text{ref},i}),
\end{equation}
where ${\bf K}_{\text{ref}}$ and ${\bf K}_{i}$ are the intrinsic parameters for the reference image $I_\text{ref}$ and the source image $I_{i}$ respectively, ${\bf R }_{\text{ref},i}$ is the relative rotation matrix, ${\bf t}_{\text{ref},i}$ is the relative translation matrix, and $d_j$ denotes the $j$-th sampled depth hypothesis. Note that, we uniformly sample depth hypotheses in the inverse depth space, i.e., $d_{j}=(\frac{1}{d_\text{min}}-(\frac{1}{d_\text{min}}-\frac{1}{d_\text{max}})\frac{j}{D-1})^{-1}, j\in\{0,\cdots,D-1\}$, where $[d_\text{min},d_\text{max}]$ is the depth interval for the reference image and $D$ is total sampling number of depth hypotheses. This allows our network to be applicable to complex and large-scale scenes~\cite{Yao2019RMVSNet,Xu2020Learning}. Once the projected point in $I_i$ is known, we use the differential bilinear interpolation~\cite{Jaderberg2015Spatial} to warp the deep features of $I_i$ into the coordinate of the reference image. Then a multi-channel similarity map between the warped source image feature and the reference image feature is computed by group-wise correlation~\cite{Guo2019Group}. According to $D$ sampling depth hypotheses, 
we compute $D$ similarity maps. By packing these similarity maps, 
the two-view cost volume of the source image $I_i$, $C_{\text{ref},i}$ of size $G\times \frac{H}{4} \times \frac{W}{4}\times D$, can be constructed, where $G=8$ is the channel number of similarity maps.  

\textbf{Pixelwise visibility-aware cost volume aggregation.} 
After obtaining the two-view cost volume $C_{\text{ref},i}$, we hope to leverage it to regress the visibility map for the source image $I_i$. This is possible because the two-view cost volume $C_{\text{ref},i}$ encodes the confidence of different sampling depths~\cite{Hu2012Quantitative,Poggi2017Quantitative,Kim2018Unified,Kim2019LAF}. The network architecture of our pixelwise visibility network is illustrated in Fig.~\ref{network} (b). Since the distribution of two-view cost volumes is often non-discriminative as mentioned in~\cite{Seki2016Patch,Fu2018Learning}, we first apply a 3D U-Net~\cite{Ronneberger2015UNet} to modulate cost volumes. Our 3D U-Net uses a three-scale encoder-decoder structure to increase the receptive field. Except for the last convolution layer that produces one-channel feature followed by the sigmoid activation function, other convolution layers are followed by a batch-normalization (BN) layer and a rectified linear unit (ReLU). Afterwards, we define the visibility map $V_i$ for the source image $I_i$ as
\begin{equation}
V_i({\bf p})=\text{max}\{{\bf P}_v(j,{\bf p})|j=0,\cdots,D-1\},
\end{equation}
where ${\bf P}_v(j,{\bf p})$ is the probability estimation for pixel ${\bf p}$ at the $j$-th sampling depth value. 
In order to further eliminate the influence of unrelated source images, we deactivate the images whose visibility probability is below a certain threshold. Then, the visibility map of each source image is modified as
\begin{equation}\label{eq:pt}
V_i'({\bf p})=
\begin{cases}
V_i({\bf p}), &\text{if $V_i({\bf p})>\tau$;} \\
0, &\text{otherwise.}
\end{cases}
\end{equation}
where $\tau=0.05$ is a threshold that controls the activation of source images. The above equation is similar to ReLU and our overall network can be trained end-to-end with back-propagation. In this way, the final aggregated cost volume is computed by
\begin{equation}
C_\text{agg}({\bf p})=\frac{\sum_{i=0}^{N-1}V_i'({\bf p})\cdot C_{\text{ref},i}({\bf p})}{\sum_{i=0}^{N-1}V_i'({\bf p})}.
\end{equation}
The final aggregated cost volume is the sum of each two-view cost volume weighted by its visibility map. 
This definition differs from all previous multi-view similarity measures~\cite{Hartmann2017Learned,Kar2017Learning,Yao2018MVSNet,Huang2018DeepMVS,Xu2020Learning} because it considers the pixelwise visibility information of each neighboring view. This not only reduces the influence of noise but also lessens the regularization burden of cost volume filtering. 

\subsection{Cost volume filtering and depth inference}

In oder to further aggregate context across the spatial domain as well as the depth domain, we apply 3D CNNs with stacked regularization modules to filter the aggregated cost volume. In particular, similar to \cite{Xu2020Learning}, our cost volume filtering module consists of a 3D ResNet~\cite{He2016Deep} and two 3D U-Nets. 
To adapt to large-scale scene reconstruction and achieve sub-pixel depth estimation, we employ the inverse depth regression in \cite{Xu2020Learning} to obtain the depth prediction,
\begin{equation}
D({\bf p})=((\frac{1}{d_\text{min}}-\frac{1}{d_\text{max}})\frac{\sum_{j=0}^{D-1}j\cdot {\bf P}(j,{\bf p})}{D-1}+\frac{1}{d_\text{max}})^{-1},
\end{equation}
where $\sum_{j=0}^{D-1}j\cdot {\bf P}(j,{\bf p})$ is the predicted ordinal and ${\bf P}(j,{\bf p})$ is the probability volume that has been normalized by the softmax operation  along the depth direction. For three regularization modules, our network produces three depth predictions, $D_\text{pred}^0$, $D_\text{pred}^1$ and $D_\text{pred}^2$. We adopt the $L_1$ loss function to train our network. The loss function for low-resolution prediction is defined as
\begin{equation}\label{eq:loss}
L=\sum_{l=0}^{2}\sum_{{\bf p}\in\Phi_0}\lambda_l\cdot||D_\text{pred}^l({\bf p})-D_\text{gt}^0({\bf p})||,
\end{equation}
where $\Phi_0$ denotes the set of valid ground truth pixels at low-resolution and $D_\text{gt}^0$ is the ground truth depth map at low-resolution. $\lambda_l$ is a scalar controlling the importance of different terms.

\subsection{Extension on high-resolution estimation}

High-quality dense 3D reconstruction requires accurate high-resolution depth map estimation. There have been some efforts to use refined depth sampling to reconstruct thin cost volumes for high-resolution depth map estimation~\cite{Cheng2020UCSNet,Yang2020CVPMVSNet,Gu2019Cas}. However, they all inherit the multi-view similarity measure from \cite{Yao2018MVSNet} to build thin cost volumes without considering the visibility information of neighboring views. To alleviate this problem in the high-resolution estimation, we directly leverage the low-resolution visibility estimation to help the thin cost volume construction. Specifically, after obtaining the predicted ordinal and probability volume at one stage, we adopt the variance-based uncertainty estimation in \cite{Cheng2020UCSNet} to compute the depth sampling range at the next stage. The visibility maps at the previous stage are directly upsampled to the current stage to compute visibility-aware thin cost volumes. Then a simple 3D U-Net is applied to obtain the probability volume, which is used to infer depth maps by inverse depth regression. The same process is iterated until the depth map reaches the original image resolution. The training loss for high-resolution prediction is defined as
\begin{equation}
L_\text{HiRes}=L+\sum_{{\bf p}\in\Phi_1}\lambda_3\cdot||D_\text{pred}^3({\bf p})-D_\text{gt}^1({\bf p})||+\sum_{{\bf p}\in\Phi_2}\lambda_4\cdot||D_\text{pred}^4({\bf p})-D_\text{gt}^2({\bf p})||,
\end{equation}
where $\Phi_1$ and $\Phi_2$ denote the set of valid ground truth pixels at the second and third stages, $D_\text{pred}^3$ and $D_\text{pred}^4$ are the predicted depth maps at the second and third stages and $D_\text{gt}^1$ and $D_\text{gt}^2$ are their corresponding ground truth depth. $\lambda_3$ and $\lambda_4$ are scalars to balance the weights of different terms.

\subsection{Anti-noise training strategy}

Without considering pixelwise visibility information, all the previous learning-based methods~\cite{Yao2019RMVSNet,Chen2019Point,Xu2020Learning,Cheng2020UCSNet} follow MVSNet~\cite{Yao2018MVSNet} to select the best two neighboring views via global view selection for model training. However, the two selected views have such strong visibility association with the reference image that only a few negative samples of visibility information participate in model training. The extreme imbalance between positives and negatives prevents our method from fully exploiting the potential of the pixelwise visibility network. 
To alleviate this problem, we propose an anti-noise training strategy (AN) that introduces disturbing views. Specifically, we adopt the method in MVSNet to compute global view selection scores of neighboring views. Then, we choose the best two views and the worst two views to train our model. This training strategy introduces more negative samples, making our network more robust to unrelated views.

\section{Experiments}

In this section, we first describe datasets, evaluation metrics and implementation details. Then we validate the effectiveness of our proposed pixelwise visibility network and anti-noise training strategy. Comparisons with several state-of-the-art methods are presented at last.

\textbf{Datasets.} We test our method on DTU dataset~\cite{Aanes2016Large}, Tanks and Temples dataset~\cite{Knapitsch2017TTB} and ETH3D high-res benchmark~\cite{Schops2017Multi}. DTU dataset is captured in the indoor controlled environment. It only contains object-centric scenes with  video sequences as input. Tanks and Temples dataset is also with video sequences as input and provides both indoor scenes and outdoor environments. It is further divided into Intermediate dataset and Advanced dataset.
ETH3D high-res benchmark is a dataset with strong viewpoint variations.

\textbf{Evaluation metrics.} The quality of depth prediction is evaluated by the commonly used mean absolute depth error (MAE). For point cloud evaluation, the accuracy and completeness of the distance metric are adopted for DTU dataset while the accuracy and completeness of the percentage metric for the other two datasets. The overall score of the distance metric is the mean of the accuracy and completeness while their $F_1$ score measures the overall score of the percentage metric.
 
 \textbf{Implementation details.} Following \cite{Ji2017Surfacenet,Yao2018MVSNet}, we split the DTU dataset~\cite{Aanes2016Large} into training, validation and evaluation set. We train our network on the training set. Poisson surface reconstruction~\cite{Kazhdan2013SPS} is leveraged to generate the ground truth depth maps at three resolutions, $\frac{H}{4}\times\frac{W}{4}$, $\frac{H}{2}\times\frac{W}{2}$ and $H\times W$. During the training, we set $H\times W=512\times640$ and $D=192$. The weights for balancing different loss terms are set to $\lambda_0=\lambda_1=0.5$, $\lambda_2=\lambda_3=\lambda_4=0.7$. Our global view selection includes $20$ neighboring views. Our network is implemented using PyTorch~\cite{NIPS2019PyTorch}. We use RMSprop optimizer to train our network on two NVIDIA GTX 1080Ti GPUs. The initial learning rate is set to $0.001$.
 
\subsection{Verification of the pixelwise visibility network and the anti-noise training strategy}

\begin{wrapfigure}{r}{5.3cm}
	\setlength{\abovecaptionskip}{0pt}
	\setlength{\belowcaptionskip}{0pt}
	\centering
	\includegraphics[width=0.365\textwidth]{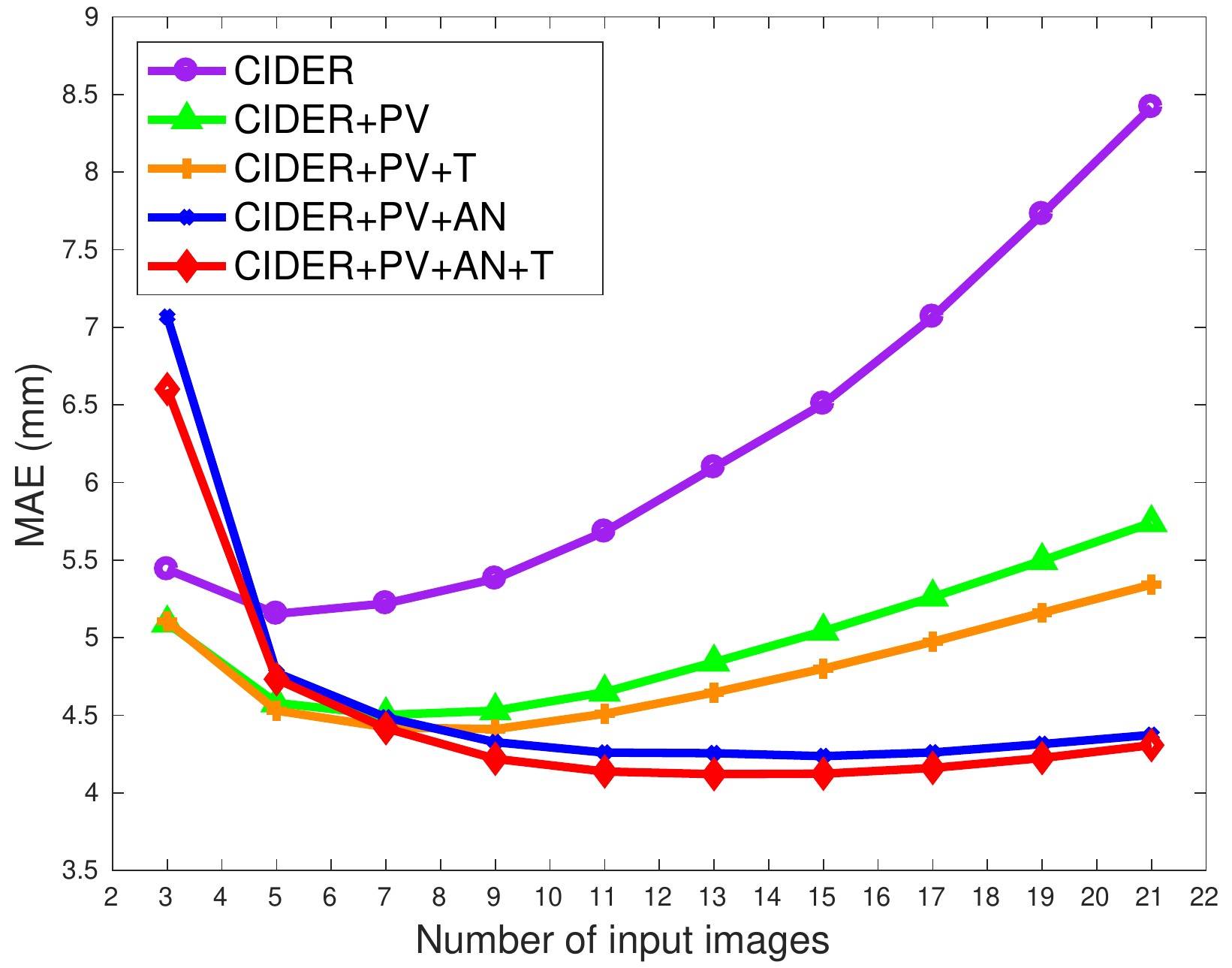}
	\caption{The prediction error of different models for varying input numbers.}
	\label{modelplot}
\end{wrapfigure}

To validate the performance of our proposed network, we conduct experiments on DTU validation set using our low-resolution model, i.e., only using Eq.~\ref{eq:loss} to train our model. 
We first verify the pixelwise visibility network and then illustrate the goodness of the anti-noise training strategy. 

\textbf{Pixelwise visibility network (PV).} We first test the pixelwise visibility network by setting the visibility map of each source image to fill with the scalar value $1$. This makes our method turn into CIDER~\cite{Xu2020Learning}, where each source image is treated equally in the multi-view cost volume aggregation. For a fair comparison, we only use the best two neighboring views to train CIDER and CIDER with PV. Here probability truncation (T) in Eq.~\ref{eq:pt} will not be used. 

We plot the MAE of different models with different view numbers in Fig.~\ref{modelplot} . CIDER achieves its best at  $5$ input views with an error of $5.154$ (the purple curve). But its performance degrades dramatically when the input views increase. 
It shows that if not considering visibility estimation, more input views will bring more noise from unrelated views, which will seriously degrade the performance of the learning-based methods. 
Therefore, existing learning-based methods cannot use more input views like traditional geometric methods, which prevents them from being applied in practice. 
Comparatively, CIDER+PV (the green curve) gets $4.580$ MAE at $5$ input views and reaches its best  at $7$ input views ($4.503$), much better than CIDER.
This shows that even for the datasets with slight viewpoint changes, visibility estimation is still crucial. 
With the increase of input views, the results of CIDER+PV also deteriorate, but the degradation is much less severe than CIDER. 
This is because the manifest imbalance between positives and negatives in the original training strategy makes PV less distinguishable to unrelated views. 
To confirm this, we directly apply the probability truncation to test CIDER+PV (the orange curve). Its MAE is further reduced and the minimal MAE is $4.411$. This shows it is useful to further lower the visibility probability of unrelated views. 
Since the original training strategy uses the best two neighboring views, their pixelwise visibility probability is relatively high in most cases. This makes PV tend to only fit the visibility of high probability during the training, resulting in its low discrimination to unrelated views. Therefore, with the increase of input views, the probability sum of unrelated views will dominate in the aggregated cost volume, making the performance degrade.  

\begin{figure}[t]
	\setlength{\abovecaptionskip}{0pt}
	\setlength{\belowcaptionskip}{0pt}
	\centering
	\includegraphics[width=0.99\textwidth]{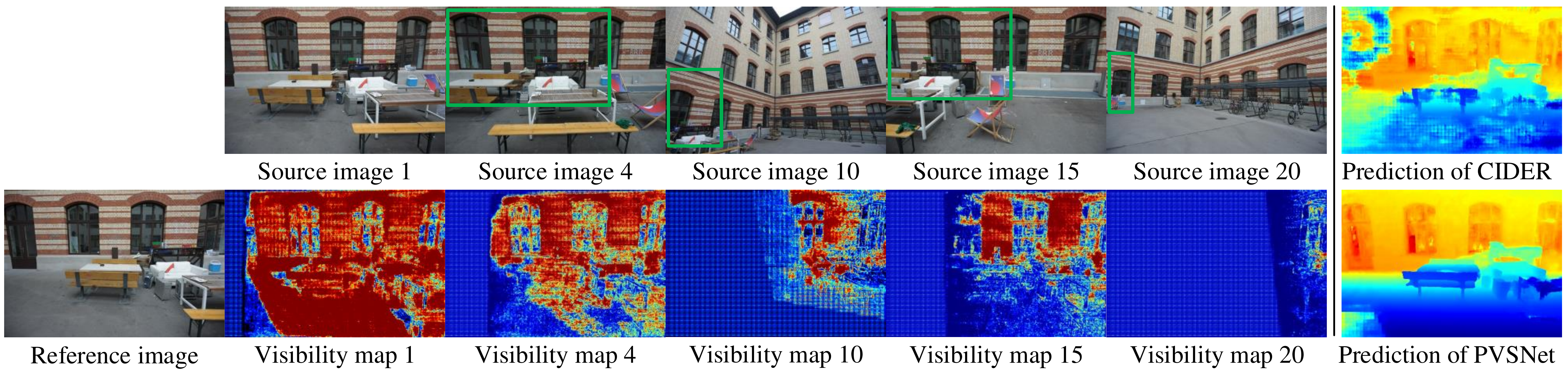}
	\caption{Left: Visibility maps show the visible areas in the reference image with respect to source images. Right: Predicted depth map comparison between CIDER and our proposed PVSNet. }\label{visualization}
\end{figure}   

\begin{table}[t]
	\begin{minipage}[t]{0.51\textwidth}
		\caption{Quantitative results on the DTU evaluation set using the distance metric ($mm$). LR means low-resolution while HR means high-resolution.}
		\centering
		\footnotesize
		\begin{tabular}{ccccc}
			\toprule
			& Method & Acc. & Comp. & Overall \\ 
			\midrule
			\multirow{4}{0.02cm}{\rotatebox{90}{Traditional}} & Camp~\cite{Camp2008Using} & 0.835 & 0.554 & 0.695 \\ 
			& Furu~\cite{Furukawa2010Accurate} & 0.613 & 0.941 & 0.777 \\
			& Tola~\cite{Tola2012Efficient} & 0.342 & 1.190 & 0.766 \\
			& Gipuma~\cite{Galliani2015Massively} & \bf{0.283} & 0.873 & 0.578 \\
			\midrule
			\multirow{5}{0.02cm}{\rotatebox{90}{LR Learning}} 
			& MVSNet~\cite{Yao2018MVSNet} & 0.396 & 0.527 & 0.462 \\
			& R-MVSNet~\cite{Yao2019RMVSNet} & 0.383 & 0.452 & 0.417 \\
			& CIDER~\cite{Xu2020Learning} & 0.417 & 0.437 & 0.427 \\
			& MVSCRF~\cite{Xue2019MVSCRF} & 0.371 & 0.426 & 0.398 \\
			& Ours & 0.408 & \bf{0.393} & 0.4001 \\
			\midrule
			\multirow{5}{0.02cm}{\rotatebox{90}{HR Learning}} 
			& Point-MVSNet~\cite{Chen2019Point} & 0.342 & 0.411 & 0.376 \\
			& CasMVSNet~\cite{Gu2019Cas} & 0.325 & 0.385 & 0.355 \\
			& CVP-MVSNet~\cite{Yang2020CVPMVSNet} & 0.296 & 0.406 & 0.351 \\
			& UCSNet~\cite{Cheng2020UCSNet} & 0.330 & 0.372 & 0.351 \\
			& Ours & 0.337 & \bf{0.315} & \bf{0.326} \\ 
			\bottomrule
		\end{tabular}
		\label{tab:dtuevaluation}
	\end{minipage}
	\hfill
	\begin{minipage}[t]{0.48\textwidth}
		\caption{Quantitative results on Tanks and Temples dataset using percentage metric ($\%$). All the values including ours are from the website~\cite{Knapitsch2017}.}
		\centering
		\footnotesize
		\begin{tabular}{ccccc}
			\toprule
			& Method & Acc. & Comp. & $F_1$ \\ 
			\midrule
			\multirow{10}{0.02cm}{\rotatebox{90}{Intermediate}} & MVSNet~\cite{Yao2018MVSNet} & 40.23 & 49.70 & 43.48 \\ 
			& R-MVSNet~\cite{Yao2019RMVSNet} & 43.74 & 57.60 & 48.60 \\
			& MVSCRF~\cite{Xue2019MVSCRF} & 41.99 & 52.58 & 45.73 \\
			& CIDER~\cite{Xu2020Learning} & 42.79 & 55.21 & 46.76 \\
			& P-MVSNet~\cite{Luo2019P} & 49.93 & 63.82 & 55.62 \\
			& Point-MVSNet~\cite{Chen2019Point} & 41.27 & 60.13 & 48.27 \\
			& CasMVSNet~\cite{Gu2019Cas} & 47.62 & \bf{74.01} & 56.84 \\
			& CVP-MVSNet~\cite{Yang2020CVPMVSNet} & 51.41 & 60.19 & 54.03 \\
			& UCSNet~\cite{Cheng2020UCSNet} & 46.66 & 70.34 & 54.83 \\
			& Ours & \bf{53.71} & 63.88 & \bf{56.88} \\
			\midrule
			\multirow{4}{0.02cm}{\rotatebox{90}{Advanced}} & R-MVSNet~\cite{Yao2019RMVSNet} & \bf{31.47} & 22.05 & 24.91 \\
			& CIDER~\cite{Xu2020Learning} & 26.64 & 21.27 & 23.12 \\
			& CasMVSNet~\cite{Gu2019Cas} & 29.68 & 35.24 & 31.12 \\
			& Ours & 29.43 & \bf{41.17} & \bf{33.46} \\
			\bottomrule
		\end{tabular}
		\label{tab:t2evaluation}
	\end{minipage}
\end{table} 

\textbf{Anti-noise training strategy (AN).} To make the pixelwise visibility network more distinguishable to unrelated views, we retrain CIDER+PV by anti-noise training strategy (AN). In contrast to the original training strategy, AN introduces the two worst neighboring views during the training.
With the AN strategy, CIDER+PV gets better and better performance with the increase of input views and reaches its best with $4.237$ MAE at $15$ input views (cf. the blue curve in Fig.~\ref{modelplot}).  
Moveover, even if the input views are increased to $21$, CIDER+PV+AN is still not degraded obviously. 
This indicates that CIDER+PV+AN can not only choose good views by assigning them higher probability, but also remove bad views by lowering their probability. 
When we further introduce the probability truncation to train CIDER+PV+AN, the results (the red curve) show its performance improves slightly and reach the best with $4.121$ MAE at $13$ input views. 
This also demonstrates our proposed AN can distinguish the unrelated views well.      
Note that, when the input views are less than $7$, CIDER+PV+AN is not better than CIDER+PV. But after that CIDER+PV+AN becomes better and better while CIDER+PV is worse and worse.
This means that the previous training strategy makes networks depend too much on the cost volume filtering while AN can make good use of multi-view information. 
Thus, when the input views are more than $7$, the cost volume filtering is not enough to counter the noise from unrelated views. In contrast, our proposed AN allows PV to distinguish unrelated views very well and eliminate their influence before the cost volume filtering, which greatly improves the performance.

\subsection{Visualization analysis of the visibility map}

So far, we have demonstrated that our overall PVSNet is able to eliminate the influence of unrelated views to improve the performance on DTU dataset. 
We further present the visualization analysis of visibility map estimation on ETH3D high-res benchmark  with strong viewpoint changes in Fig.~\ref{visualization}.
As can be seen, these source images contain strong variations in viewpoint. If they are treated equally in the cost volume aggregation, the depth prediction of the reference image will be quite inaccurate (cf. the depth prediction of CIDER in Fig.~\ref{visualization}). With the help of our proposed pixelwise visibility network, our method can reasonably identify the visible areas in the reference image with respect to different source image. For example, our proposed pixelwise visibility network indicates that most areas of the reference image are visible in source image $1$ and source image $4$. As for visibility map $10$, $15$ and $20$, they indicate that only partial areas of the reference image are visible in source image $10$, $15$ and $20$, which are denoted as the green boxes in Fig.~\ref{visualization}.  
By only including these visible areas in the cost volume aggregation, our method can accurately recover the depth information of the reference image (cf. the depth prediction of PVSNet in Fig.~\ref{visualization}).

\subsection{Comparisons with the state-of-the-art}

In this section, we compare our method with stat-of-the-art methods on different datasets. We use the filtering and fusion method in MVSNet~\cite{Yao2018MVSNet} to produce point clouds. To leverage the multi-view information as much as possible, we set $N=11$ for all datasets. Note that, we use the model trained on DTU training set to test all datasets without any fine-tuning. 

\textbf{DTU dataset.} As illustrated in Table~\ref{tab:dtuevaluation}, the high resolution estimation of PVSNet outperforms all state-of-the-art methods in terms of completenss and overall scores. Besides, the low resolution version of PVSNet achieves the best completeness and almost the best overall score among the given learning-based methods with low resolution output.

\begin{wraptable}{r}{5.2cm}
	\setlength{\abovecaptionskip}{0pt}
	\setlength{\belowcaptionskip}{0pt}
	\caption{$F_1$ score (in $\%$) comparisons
		of point clouds on ETH3D high-res benchmark at evaluation
		threshold $2cm$. Colmap (low) means that input images are resized to $H\times W=1280\times1920$. All the values including ours are from the website~\cite{Schops2017ETH3D}.}
	\centering
	\small
	\begin{tabular}{ccc}
		\toprule
		Method & train & test  \\
		\midrule
		MVE~\cite{Fuhrmann2015MVE} & 20.47 & 30.37 \\
		Gipuma~\cite{Galliani2015Massively} & 36.38 & 45.18 \\
		PMVS~\cite{Furukawa2010Accurate} & 46.06 & 44.16 \\
		Colmap (low)~\cite{Schonberger2016Pixelwise} & 60.54 & - \\
		Colmap~\cite{Schonberger2016Pixelwise} & 67.66 & 73.01 \\
		Ours & 67.48 & 72.08 \\
		\bottomrule 
	\end{tabular}
	\label{tab:eth3devaluation}
\end{wraptable}

\textbf{Tanks and Temples dataset.} As shown in Table~\ref{tab:t2evaluation}, our method achieves the best results on both Intermediate dataset and Advanced dataset. This demonstrates that our pixelwise visibility network also improves the performance on datasets with video sequences as input. Note that, since the Advanced dataset contains stronger viewpoint variations, our method has a greater performance improvement on this dataset. 

\textbf{ETH3D high-res benchmark.} The image resolution of this benchmark is very high, reaching about $H\times W=4000\times6000$. We set image size $H\times W=1280\times1920$ for this benchmark due to the GPU memory limitation. Table~\ref{tab:eth3devaluation} summarizes the $F_1$ scores of different methods. As aforementioned, this benchmark contains strong viewpoint variations. Without considering visibility estimation, existing learning-based methods have all failed on this benchmark. Our method is the first learning-based method that can be truly evaluated on this benchmark. We can see that our method has surpassed some traditional methods, such as MVE~\cite{Fuhrmann2015MVE}, Gipuma~\cite{Galliani2015Massively} and PMVS~\cite{Furukawa2010Accurate}. Moreover, our method is on par with Colmap~\cite{Schonberger2016Pixelwise}. Note that, the resolution of the images tested in our network, $1280\times1920$, is lower than Colmap~\cite{Schonberger2016Pixelwise}, $2130\times3200$\footnote{In fact, Colmap's evaluation results on ETH3D high-res benchmark are obtained by resizing the high-res images to $2130\times3200$.}. When we resize the input high-res images to $1280\times1920$ and rerun Colmap, the results in Table~\ref{tab:eth3devaluation} show that the $F_1$ scores of our method are better than Colmap with low resolution input. This further demonstrates that our method can be generalized well to this challenging dataset. This is impossible for the previous learning-based methods.

\section{Conclusion}

In this work, we propose the Pixelwise Visibility-aware multi-view Stereo Network (PVSNet) for robust dense 3D reconstruction. Different from most of existing learning-based methods that treat each neighboring view equally, our network learns to estimate the pixelwise visibility information of neighboring views to construct a weighted aggregated cost volume. This helps to eliminate the influence of unrelated neighboring views in advance. In order to make the pixelwise visibility network more discriminative to the unrelated views, we present a novel training strategy by introducing disturbing views. In this way, our method can be generalized well to different types of datasets, including the challenging ETH3D high-res benchmark. Moreover, extensive experiments demonstrate the superiority of PVSNet on dense 3D reconstruction. Since our proposed pixelwise visibility network is independent of different baseline models, its greater potential will be explored when combined with better baseline models  in the future.

\bibliographystyle{unsrt}
\bibliography{refs}

\begin{thebibliography}{10}

\bibitem{Seitz2006Comparison}
Steven~M. Seitz, Brian Curless, James Diebel, Daniel Scharstein, and Richard
  Szeliski.
\newblock A comparison and evaluation of multi-view stereo reconstruction
  algorithms.
\newblock In {\em Proceedings of the IEEE Computer Society Conference on
  Computer Vision and Pattern Recognition}, page 519–528, 2006.

\bibitem{Furukawa2015MST}
Yasutaka Furukawa and Carlos Hern\'{a}ndez.
\newblock Multi-view stereo: A tutorial.
\newblock {\em Found. Trends. Comput. Graph. Vis.}, 9(1-2):1--148, 2015.

\bibitem{Furukawa2010Accurate}
Y.~Furukawa and J.~Ponce.
\newblock Accurate, dense, and robust multiview stereopsis.
\newblock {\em IEEE Transactions on Pattern Analysis and Machine Intelligence},
  32(8):1362--1376, 2010.

\bibitem{Schonberger2016Pixelwise}
Johannes~L. Sch{\"o}nberger, Enliang Zheng, Jan-Michael Frahm, and Marc
  Pollefeys.
\newblock Pixelwise view selection for unstructured multi-view stereo.
\newblock In {\em Proceedings of the European Conference on Computer Vision},
  pages 501--518, 2016.

\bibitem{Yao2018MVSNet}
Yao Yao, Zixin Luo, Shiwei Li, Tian Fang, and Long Quan.
\newblock Mvsnet: Depth inference for unstructured multi-view stereo.
\newblock In {\em Proceedings of the European Conference on Computer Vision},
  pages 767--783, 2018.

\bibitem{Xu2019Multi}
Qingshan Xu and Wenbing Tao.
\newblock Multi-scale geometric consistency guided multi-view stereo.
\newblock In {\em Proceedings of the IEEE Conference on Computer Vision and
  Pattern Recognition}, pages 5483--5492, 2019.

\bibitem{Xu2020Planar}
Qingshan Xu and Wenbing Tao.
\newblock Planar prior assisted patchmatch multi-view stereo.
\newblock In {\em Proceedings of the AAAI Conference on Artificial
  Intelligence}, 2020.

\bibitem{Yao2019RMVSNet}
Yao Yao, Zixin Luo, Shiwei Li, Tianwei Shen, Tian Fang, and Long Quan.
\newblock Recurrent mvsnet for high-resolution multi-view stereo depth
  inference.
\newblock In {\em Proceedings of the IEEE Conference on Computer Vision and
  Pattern Recognition}, pages 5525--5534, 2019.

\bibitem{Chen2019Point}
Rui Chen, Songfang Han, Jing Xu, and Hao Su.
\newblock Point-based multi-view stereo network.
\newblock In {\em Proceedings of the IEEE International Conference on Computer
  Vision}, pages 1538--1547, 2019.

\bibitem{Xu2020Learning}
Qingshan Xu and Wenbing Tao.
\newblock Learning inverse depth regression for multi-view stereo with
  correlation cost volume.
\newblock In {\em Proceedings of the AAAI Conference on Artificial
  Intelligence}, 2020.

\bibitem{Cheng2020UCSNet}
Shuo Cheng, Zexiang Xu, Shilin Zhu, Zhuwen Li, Li~Erran Li, Ravi Ramamoorthi,
  and Hao Su.
\newblock Deep stereo using adaptive thin volume representation with
  uncertainty awareness, 2020.

\bibitem{Knapitsch2017TTB}
Arno Knapitsch, Jaesik Park, Qian-Yi Zhou, and Vladlen Koltun.
\newblock Tanks and temples: Benchmarking large-scale scene reconstruction.
\newblock {\em ACM Trans. Graph.}, 36(4):78:1--78:13, 2017.

\bibitem{Schops2017Multi}
T.~Sch{\"o}ps, J.~L. Sch{\"o}nberger, S.~Galliani, T.~Sattler, K.~Schindler,
  M.~Pollefeys, and A.~Geiger.
\newblock A multi-view stereo benchmark with high-resolution images and
  multi-camera videos.
\newblock In {\em Proceedings of the IEEE Conference on Computer Vision and
  Pattern Recognition}, pages 2538--2547, 2017.

\bibitem{Kang2001Handling}
Sing~Bing Kang, R.~Szeliski, and Jinxiang Chai.
\newblock Handling occlusions in dense multi-view stereo.
\newblock In {\em Proceedings of the IEEE Conference on Computer Vision and
  Pattern Recognition}, volume~1, pages I--I, 2001.

\bibitem{Galliani2015Massively}
S.~Galliani, K.~Lasinger, and K.~Schindler.
\newblock Massively parallel multiview stereopsis by surface normal diffusion.
\newblock In {\em Proceedings of the IEEE International Conference on Computer
  Vision}, pages 873--881, 2015.

\bibitem{Goesele2006Multi}
M.~{Goesele}, B.~{Curless}, and S.~M. {Seitz}.
\newblock Multi-view stereo revisited.
\newblock In {\em Proceedings of the IEEE Conference on Computer Vision and
  Pattern Recognition}, volume~2, pages 2402--2409, June 2006.

\bibitem{Zheng2014PatchMatch}
E.~Zheng, E.~Dunn, V.~Jojic, and J.~M. Frahm.
\newblock Patchmatch based joint view selection and depthmap estimation.
\newblock In {\em Proceedings of the IEEE Conference on Computer Vision and
  Pattern Recognition}, pages 1510--1517, 2014.

\bibitem{Hartmann2017Learned}
W.~{Hartmann}, S.~{Galliani}, M.~{Havlena}, L.~V. {Gool}, and K.~{Schindler}.
\newblock Learned multi-patch similarity.
\newblock In {\em Proceedings of the IEEE International Conference on Computer
  Vision}, pages 1595--1603, 2017.

\bibitem{Ji2017Surfacenet}
Mengqi Ji, Juergen Gall, Haitian Zheng, Yebin Liu, and Lu~Fang.
\newblock Surfacenet: An end-to-end 3d neural network for multiview stereopsis.
\newblock In {\em Proceedings of the IEEE International Conference on Computer
  Vision}, pages 2307--2315, 2017.

\bibitem{Kar2017Learning}
Abhishek Kar, Christian H\"{a}ne, and Jitendra Malik.
\newblock Learning a multi-view stereo machine.
\newblock In {\em Advances in Neural Information Processing Systems}, pages
  365--376. 2017.

\bibitem{Huang2018DeepMVS}
P.~{Huang}, K.~{Matzen}, J.~{Kopf}, N.~{Ahuja}, and J.~{Huang}.
\newblock Deepmvs: Learning multi-view stereopsis.
\newblock In {\em Proceedings of the IEEE Conference on Computer Vision and
  Pattern Recognition}, pages 2821--2830, June 2018.

\bibitem{Xue2019MVSCRF}
Youze Xue, Jiansheng Chen, Weitao Wan, Yiqing Huang, Cheng Yu, Tianpeng Li, and
  Jiayu Bao.
\newblock Mvscrf: Learning multi-view stereo with conditional random fields.
\newblock In {\em Proceedings of the IEEE International Conference on Computer
  Vision}, pages 4312--4321, 2019.

\bibitem{Luo2019P}
Keyang Luo, Tao Guan, Lili Ju, Haipeng Huang, and Yawei Luo.
\newblock P-mvsnet: Learning patch-wise matching confidence aggregation for
  multi-view stereo.
\newblock In {\em Proceedings of the IEEE International Conference on Computer
  Vision}, pages 10452--10461, 2019.

\bibitem{Im2019Dpsnet}
Sunghoon Im, Hae-Gon Jeon, Stephen Lin, and In~So Kweon.
\newblock Dpsnet: End-to-end deep plane sweep stereo.
\newblock In {\em Proceedings of the International Conference on Learning
  Representations}, 2019.

\bibitem{Collins1996Space}
R.~T. {Collins}.
\newblock A space-sweep approach to true multi-image matching.
\newblock In {\em Proceedings of the IEEE Conference on Computer Vision and
  Pattern Recognition}, pages 358--363, 1996.

\bibitem{Jaderberg2015Spatial}
Max Jaderberg, Karen Simonyan, Andrew Zisserman, and koray kavukcuoglu.
\newblock Spatial transformer networks.
\newblock In {\em Advances in Neural Information Processing Systems}, pages
  2017--2025. 2015.

\bibitem{Guo2019Group}
Xiaoyang Guo, Kai Yang, Wukui Yang, Xiaogang Wang, and Hongsheng Li.
\newblock Group-wise correlation stereo network.
\newblock In {\em Proceedings of the IEEE Conference on Computer Vision and
  Pattern Recognition}, pages 3273--3282, 2019.

\bibitem{Hu2012Quantitative}
X.~{Hu} and P.~{Mordohai}.
\newblock A quantitative evaluation of confidence measures for stereo vision.
\newblock {\em IEEE Transactions on Pattern Analysis and Machine Intelligence},
  34(11):2121--2133, 2012.

\bibitem{Poggi2017Quantitative}
M.~{Poggi}, F.~{Tosi}, and S.~{Mattoccia}.
\newblock Quantitative evaluation of confidence measures in a machine learning
  world.
\newblock In {\em Proceedings of the IEEE International Conference on Computer
  Vision}, pages 5238--5247, 2017.

\bibitem{Kim2018Unified}
S.~{Kim}, D.~{Min}, S.~{Kim}, and K.~{Sohn}.
\newblock Unified confidence estimation networks for robust stereo matching.
\newblock {\em IEEE Transactions on Image Processing}, 28(3):1299--1313, 2019.

\bibitem{Kim2019LAF}
Sunok Kim, Seungryong Kim, Dongbo Min, and Kwanghoon Sohn.
\newblock Laf-net: Locally adaptive fusion networks for stereo confidence
  estimation.
\newblock In {\em Proceedings of the IEEE Conference on Computer Vision and
  Pattern Recognition}, pages 205--214, 2019.

\bibitem{Seki2016Patch}
Akihito Seki and Marc Pollefeys.
\newblock Patch based confidence prediction for dense disparity map.
\newblock In {\em Proceedings of the British Machine Vision Conference}, pages
  23.1--23.13, 2016.

\bibitem{Fu2018Learning}
Z.~{Fu} and M.~{Ardabilian Fard}.
\newblock Learning confidence measures by multi-modal convolutional neural
  networks.
\newblock In {\em Proceedings of the IEEE Winter Conference on Applications of
  Computer Vision}, pages 1321--1330, 2018.

\bibitem{Ronneberger2015UNet}
Olaf Ronneberger, Philipp Fischer, and Thomas Brox.
\newblock U-net: Convolutional networks for biomedical image segmentation.
\newblock In {\em Medical Image Computing and Computer-Assisted Intervention --
  MICCAI 2015}, pages 234--241, 2015.

\bibitem{He2016Deep}
K.~{He}, X.~{Zhang}, S.~{Ren}, and J.~{Sun}.
\newblock Deep residual learning for image recognition.
\newblock In {\em Proceedings of the IEEE Conference on Computer Vision and
  Pattern Recognition}, pages 770--778, 2016.

\bibitem{Yang2020CVPMVSNet}
Jiayu Yang, Wei Mao, Jose~M. Alvarez, and Miaomiao Liu.
\newblock Cost volume pyramid based depth inference for multi-view stereo.
\newblock In {\em Proceedings of the IEEE Conference on Computer Vision and
  Pattern Recognition}, 2020.

\bibitem{Gu2019Cas}
Xiaodong Gu, Zhiwen Fan, Siyu Zhu, Zuozhuo Dai, Feitong Tan, and Ping Tan.
\newblock Cascade cost volume for high-resolution multi-view stereo and stereo
  matching.
\newblock In {\em Proceedings of the IEEE Conference on Computer Vision and
  Pattern Recognition}, 2020.

\bibitem{Aanes2016Large}
Henrik Aan{\ae}s, Rasmus~Ramsb{\o}l Jensen, George Vogiatzis, Engin Tola, and
  Anders~Bjorholm Dahl.
\newblock Large-scale data for multiple-view stereopsis.
\newblock {\em International Journal of Computer Vision}, 120(2):153--168,
  2016.

\bibitem{Kazhdan2013SPS}
Michael Kazhdan and Hugues Hoppe.
\newblock Screened poisson surface reconstruction.
\newblock {\em ACM Trans. Graph.}, 32(3):29:1--29:13, July 2013.

\bibitem{NIPS2019PyTorch}
Adam Paszke, Sam Gross, Francisco Massa, Adam Lerer, James Bradbury, Gregory
  Chanan, Trevor Killeen, Zeming Lin, Natalia Gimelshein, Luca Antiga, Alban
  Desmaison, Andreas Kopf, Edward Yang, Zachary DeVito, Martin Raison, Alykhan
  Tejani, Sasank Chilamkurthy, Benoit Steiner, Lu~Fang, Junjie Bai, and Soumith
  Chintala.
\newblock Pytorch: An imperative style, high-performance deep learning library.
\newblock In {\em Advances in Neural Information Processing Systems 32}, pages
  8026--8037. 2019.

\bibitem{Camp2008Using}
Neill D.~F. Campbell, George Vogiatzis, Carlos Hern{\'a}ndez, and Roberto
  Cipolla.
\newblock Using multiple hypotheses to improve depth-maps for multi-view
  stereo.
\newblock In {\em Proceedings of the European Conference on Computer Vision},
  pages 766--779, 2008.

\bibitem{Tola2012Efficient}
Engin Tola, Christoph Strecha, and Pascal Fua.
\newblock Efficient large-scale multi-view stereo for ultra high-resolution
  image sets.
\newblock {\em Machine Vision and Applications}, 23:903--920, 2012.

\bibitem{Knapitsch2017}
Arno Knapitsch, Jaesik Park, Qian-Yi Zhou, and Vladlen Koltun.
\newblock {T}anks and {T}emples {B}enchmark.
\newblock \url{https://www.tanksandtemples.org}.

\bibitem{Schops2017ETH3D}
Thomas Sch{\"o}ps, Johannes~L. Sch{\"o}nberger, Silvano Galliani, Torsten
  Sattler, Konrad Schindler, Marc Pollefeys, and Andreas Geiger.
\newblock {ETH3D} {B}enchmark.
\newblock \url{https://www.eth3d.net}.

\bibitem{Fuhrmann2015MVE}
Simon Fuhrmann, Fabian Langguth, and Michael Goesele.
\newblock Mve: A multi-view reconstruction environment.
\newblock In {\em Proceedings of the Eurographics Workshop on Graphics and
  Cultural Heritage}, page 11–18, 2014.

\end{thebibliography}

\end{document}